\documentclass[pdflatex,sn-mathphys-num,iicol]{sn-jnl}

\usepackage{graphicx}%
\usepackage{multirow}%
\usepackage{amsmath,amssymb,amsfonts}%
\usepackage{amsthm}%
\usepackage{mathrsfs}%
\usepackage[title]{appendix}%
\usepackage{xcolor}%
\usepackage{textcomp}%
\usepackage{manyfoot}%
\usepackage{booktabs}%
\usepackage{algorithm}%
\usepackage{algorithmic}%
\usepackage{array}%
\usepackage{colortbl}%
\usepackage{pifont}%
\usepackage{url}%

\definecolor{mygray}{gray}{.9}
\definecolor{myblue}{RGB}{93,80,180}
\definecolor{mygreen}{RGB}{93,173,85}
\definecolor{green}{RGB}{113,165,55}
\definecolor{blue}{RGB}{1,158,213}
\definecolor{red}{RGB}{220,10,10}
\definecolor{LightGreen}{HTML}{d4edda}
\definecolor{LightBlue}{HTML}{d1ecf1}
\definecolor{darkblue}{rgb}{0.0, 0.0, 0.55}

\begin{document}

\title[HGSQ for Real-Time Aerial Small Object Detection]{HGSQ: Heatmap-Guided Sparse Query Detector for Real-Time Aerial Small Object Detection}

\author[1]{\fnm{Yangchen} \sur{Zeng}}\email{220245765@seu.edu.cn}

\affil[1]{\orgdiv{School of Cyber Science and Engineering}, \orgname{Southeast University}, \orgaddress{\city{Nanjing}, \postcode{211189}, \country{China}}}

\abstract{Real-time aerial small object detection is an important visual signal and image processing problem, requiring a detector to preserve fine-grained localization while avoiding redundant computation on large background regions. This paper focuses on this deployment-oriented aerial/UAV setting rather than claiming a universal detector for all object detection scenarios. Existing Transformer-based detectors provide strong global modeling, but their dense query initialization and multi-layer decoder still spend substantial computation on background tokens, which is inefficient when small objects occupy only sparse image regions. To address this problem, this paper proposes \textbf{HGSQ}, a \textbf{Heatmap-Guided Sparse Query Detector} for real-time aerial small object detection. HGSQ uses a lightweight \textbf{Heatmap Budget Predictor} (HBP) to predict a foreground budget map in a single forward pass. The predicted heatmap is then used by three fixed components: \textbf{Heatmap-Guided Sparse Query Selection} (HSQS), which initializes decoder queries from high-confidence foreground positions; \textbf{Heatmap-Gated Lite Snake Convolution} (HGLSConv), which performs local shape refinement only on heatmap-activated small-object regions; and \textbf{Adaptive Query-Decoder Budgeting} (AQDB), which adjusts the query budget and decoder depth according to the estimated object density. Unlike post-hoc heatmap generation, HGSQ treats the heatmap as a real-time computation budget rather than a visualization map during deployment. Experiments on NWPU VHR-10 and VisDrone2019 show that HGSQ achieves 95.10 mAP50 on NWPU VHR-10 and 54.8 mAP50 on VisDrone2019, while reducing GFLOPs to 48.6 and running at 96.0 FPS on an RTX 4070 under our TensorRT FP16 deployment protocol.}


\maketitle

\label{sec:intro}
\begin{figure}[!t]
	\centering
	\includegraphics[width=\columnwidth]{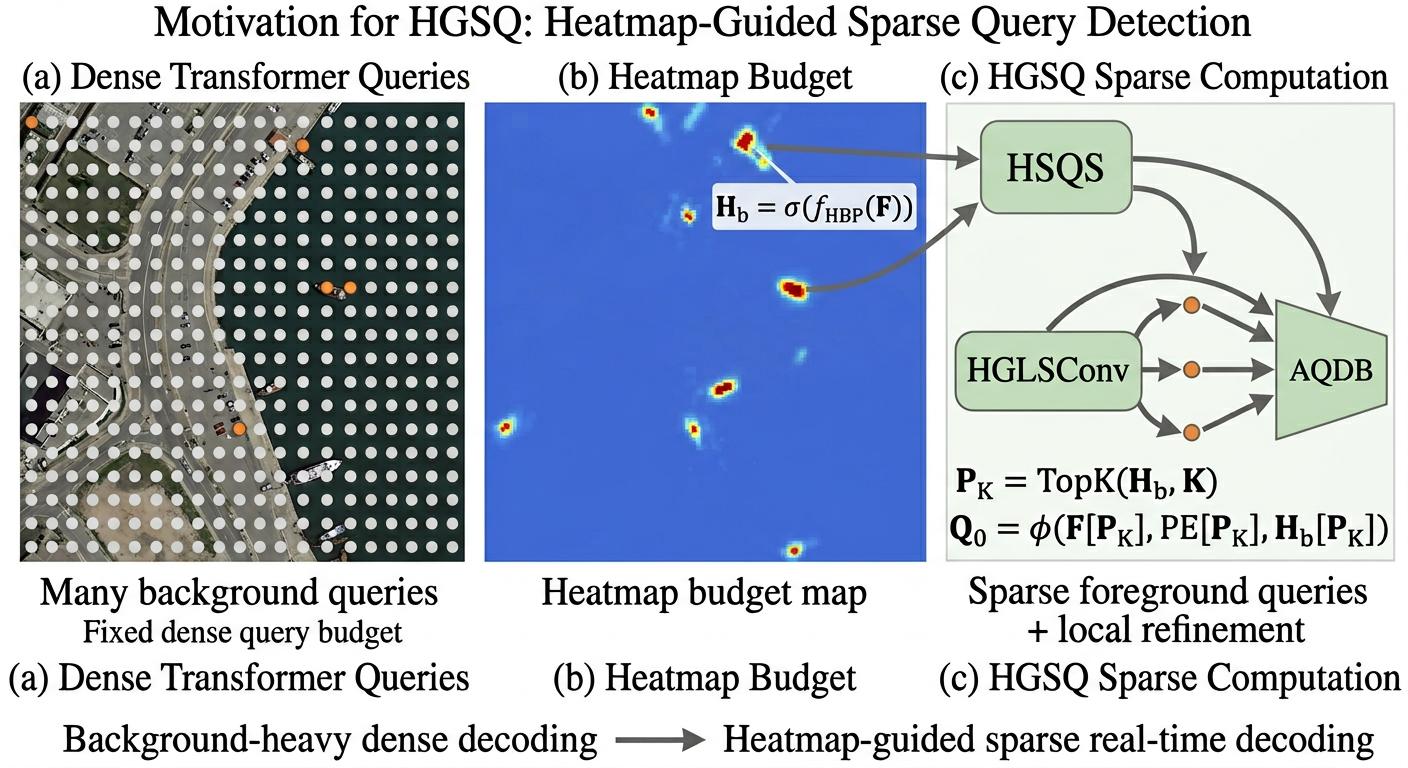}
	\caption{Motivation of HGSQ. Dense Transformer decoding allocates many queries to background regions. HGSQ predicts a heatmap budget map and selects sparse foreground queries for real-time detection.}
	\label{fig:intro_motivation}
\end{figure}

\section{Introduction}

Aerial small object detection is an important task in remote sensing and UAV perception. It is also a typical real-time image processing problem: the detector must locate dense and tiny targets from high-background-ratio images while satisfying strict latency and throughput constraints, consistent with recent Signal, Image and Video Processing studies on efficient signal detection, real-time rendering, visual recognition, segmentation, lightweight detection, and image encryption\cite{gu2026hf,jiang2026video,guo2026light,yuan2026tpe,akhtari2026comparative,fang2026infrared,zhang2025linguistic}. CNN-based real-time detectors such as the YOLO series are efficient, but their local feature aggregation may be insufficient for dense small objects in cluttered aerial scenes. Transformer-based detectors provide stronger global interaction and set prediction, yet dense object queries and repeated decoder refinement introduce redundant computation, especially when most image regions are background.

This paper narrows the scope to \textbf{real-time aerial small object detection}. Our goal is not to design a generic visualization method or a general-purpose detector for all object detection scenarios. Instead, we address a specific computation-allocation problem: in aerial images, only a small fraction of spatial positions contain small objects, but dense Transformer decoding still allocates similar computation to foreground and background regions. This mismatch limits the accuracy-speed trade-off, as illustrated in Figure~\ref{fig:intro_motivation}.

We propose \textbf{HGSQ}, a \textbf{Heatmap-Guided Sparse Query Detector}. The central idea is to use a lightweight heatmap as a \emph{real-time computation budget map}. Given feature maps extracted by the backbone, a \textbf{Heatmap Budget Predictor} (HBP) predicts a foreground heatmap in a single forward pass. The heatmap is not generated by Grad-CAM\cite{ennab2025advancing} and is not used as post-hoc interpretation. Instead, it controls where the detector should spend computation. Specifically, HGSQ uses the heatmap to select foreground queries, activate local shape refinement, and adapt the query-decoder budget. This deployment-time use distinguishes HGSQ from heatmap visualization and generic token pruning: the heatmap directly determines query construction, local refinement, and decoder budget within the detector.

HGSQ contains three fixed modules. First, \textbf{Heatmap-Guided Sparse Query Selection} (HSQS) selects top-confidence positions from the predicted heatmap and converts the corresponding image features into decoder queries. This reduces background queries before decoding. Second, \textbf{Heatmap-Gated Lite Snake Convolution} (HGLSConv) refines local features only on heatmap-activated regions. It uses lightweight axis-wise snake sampling with bounded offsets and a heatmap gate, avoiding the cost of applying deformable refinement to the whole feature map. Third, \textbf{Adaptive Query-Decoder Budgeting} (AQDB) estimates image-level object density from the heatmap and adaptively sets the query budget and decoder depth. Sparse scenes use fewer queries and shallower decoding, while dense scenes reserve more computation.

The contributions are summarized as follows:
\begin{itemize}
    \item We propose HGSQ, a heatmap-guided sparse Transformer detector for real-time aerial small object detection. HGSQ uses a single-forward heatmap budget map to allocate computation instead of using gradient-based heatmaps during inference.
    \item We design HSQS, which initializes decoder queries from heatmap-selected foreground positions and reduces redundant background decoding.
    \item We design HGLSConv, a \textbf{Heatmap-Gated Lite Snake Convolution} that performs bounded, axis-wise local refinement only in small-object candidate regions.
    \item We introduce AQDB, which adapts the query number and decoder depth according to heatmap-estimated object density, improving the accuracy-speed trade-off for real-time detection.
\end{itemize}

\section{Related Work}

\subsection{Real-time aerial small object detection}

CNN detectors such as SSD\cite{liu2016ssd}, YOLO variants\cite{redmon2018yolov3,zyc2022behavior}, Faster R-CNN\cite{ren2015faster}, and Mask R-CNN\cite{he2017mask} remain widely used because of their efficient feature extraction and mature deployment toolchains. For aerial and remote-sensing scenes, recent works improve multi-scale representation and small-object sensitivity\cite{zeng2025hmpe,zeng2026learning}. Aerial and UAV perception can also be affected by adverse illumination or weather conditions, where low contrast and degraded visibility make small objects harder to distinguish\cite{she2024mpc,she2025exploring}. These degraded-scene studies improve robust visual representation or enhancement under difficult imaging conditions, while HGSQ focuses on real-time computation allocation once aerial features are extracted. However, dense feature processing still spends substantial computation on high-background-ratio regions. HGSQ addresses this accuracy-speed conflict by allocating sparse query decoding and local refinement according to a foreground heatmap budget.

\subsection{Transformer detectors and sparse computation}

DETR\cite{carion2020end} and its variants, including Conditional DETR\cite{meng2021conditional}, DINO\cite{zhang2022dino}, Deformable DETR\cite{zhu2020deformable}, and RT-DETR\cite{zhao2024detrs}, provide strong global modeling for cluttered scenes but usually decode a fixed set of queries. Sparse computation in vision Transformers has been studied through window attention, downsampling, low-rank attention, token selection, and related sparse visual modeling\cite{zeng2026trialigngr,hong2026uv,hong2026anomaly,yu2026can,guo2026embedding,fu2026motioncraft,dong2026generated}, and region-focused detection\cite{li2024sparseformer}. Recent Transformer and local-geometry studies further improve visual representation, long-range modeling, and structure-aware feature interaction\cite{yuan2024inlier,zhang2026igasa,yuan2023egst}. Beyond visual detection, adaptive candidate selection, multimodal interest modeling, state-conditioned routing, and structured reasoning have also been explored in recommendation and knowledge-centric tasks\cite{zeng2026deepinterestgr,wang2026agent4poi,yu2026cast,fu2026neurosymactive}. Unlike offline visualization or generic token pruning, HGSQ uses a single-forward heatmap to decide which positions become decoder queries and how much decoding computation each aerial image receives.

\subsection{Local geometric refinement}

Local geometric refinement is useful for tiny aerial objects, but dense deformable convolution and deformable attention\cite{zhu2020deformable} add cost when applied to all positions. Snake-style convolution can model local structures, yet global refinement is inefficient in foreground-sparse scenes. HGLSConv therefore combines bounded axis-wise sampling with heatmap gating, so local refinement is applied mainly to high-budget small-object regions while background features remain lightweight.

\begin{figure*}[!t]
	\centering
	\includegraphics[width=0.86\textwidth,trim=0 0 0 36,clip]{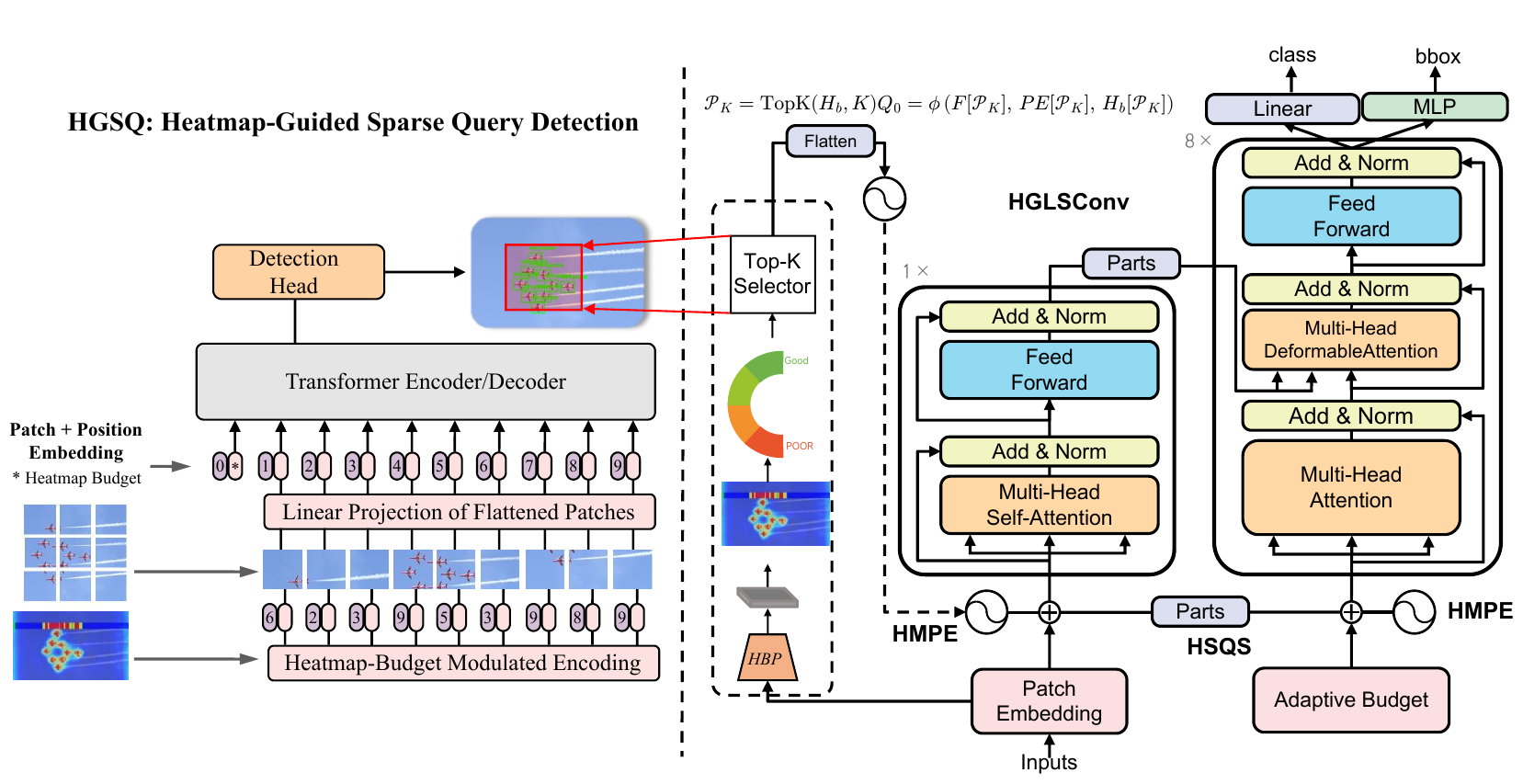}
	\caption{Overview of the proposed HGSQ framework. The Heatmap Budget Predictor (HBP) produces a foreground budget map that guides foreground-oriented query construction, heatmap-gated local refinement, and adaptive query-decoder budgeting in a single forward pass. The heatmap-guided path allocates more computation to high-budget small-object regions while reducing redundant processing of background regions.}
	\label{fig:hgsq_pipeline}
\end{figure*}

\section{Methods}

\subsection{Overview of HGSQ}

Let an input image be $I \in \mathbb{R}^{3\times H_0\times W_0}$. A backbone and neck extract multi-scale features
\begin{equation}
    \{F_l\}_{l=1}^{L} = \Phi(I), \qquad F_l \in \mathbb{R}^{C_l\times H_l\times W_l}.
\end{equation}
HGSQ applies the proposed modules mainly to the high-resolution small-object feature levels, e.g., $P_3$ and $P_4$, where tiny targets still retain spatial detail. Figure~\ref{fig:hgsq_pipeline} gives an overview of the whole pipeline. For clarity, the following equations use one feature map $F \in \mathbb{R}^{C\times H\times W}$; the same operation can be applied independently to selected feature levels.

The key design is a foreground budget heatmap $H_b \in [0,1]^{H\times W}$ predicted by a lightweight \textbf{Heatmap Budget Predictor} (HBP). HGSQ uses $H_b$ for three purposes: sparse query selection, heatmap-gated local refinement, and adaptive query-decoder budgeting. During inference, all modules are computed in a single forward pass. No Grad-CAM, backward propagation, high-order derivative, or second forward pass is used.

\subsection{Heatmap Budget Predictor}

The Heatmap Budget Predictor is a lightweight convolutional head composed of a $3\times3$ convolution, normalization and activation, a $1\times1$ projection, and a sigmoid output. The output value $H_b(i,j)$ estimates how much real-time detection computation should be allocated to spatial position $(i,j)$.

During training, the heatmap target $H^{gt}\in[0,1]^{H\times W}$ is generated from ground-truth boxes by placing Gaussian responses at object centers, with the scale proportional to the projected object size. HBP is supervised by a focal heatmap loss and trained jointly with the detector, so the heatmap learns foreground allocation from the same annotations used for detection. No offline teacher heatmap is used in our experiments. At inference time, HBP is a normal forward branch and introduces neither backward propagation nor an additional inference pass.

\subsection{Heatmap-Guided Sparse Query Selection}

Conventional Transformer detectors often decode a fixed set of learned or dense encoder queries. In aerial images, most of these queries correspond to background. HSQS replaces dense query initialization with heatmap-selected foreground queries.

Let $\Omega=\{1,\ldots,H\}\times\{1,\ldots,W\}$ be the spatial index set. Given a query budget $K$, HSQS selects
\begin{equation}
    \mathcal{P}_K = \operatorname{TopK}_{(i,j)\in\Omega}\left(H_b(i,j), K\right),
    \label{eq:topk}
\end{equation}
where $\mathcal{P}_K$ contains the $K$ positions with the highest heatmap scores. The initial decoder queries are generated by projecting the corresponding feature vectors and positional encodings:
\begin{equation}
\begin{aligned}
    q_m^{0} &= W_q F(p_m)+W_p PE(p_m) \\
            &\quad + W_h\,e(H_b(p_m)), \qquad p_m\in\mathcal{P}_K,
\end{aligned}
\label{eq:query_init}
\end{equation}
where $F(p_m)\in\mathbb{R}^{C}$ is the feature vector at $p_m$, $PE(p_m)$ is the positional encoding, $e(\cdot)$ maps a scalar heatmap score to an embedding, and $W_q$, $W_p$, and $W_h$ are learnable linear projections. The query matrix is
\begin{equation}
    Q_0=[q_1^0, q_2^0,\ldots,q_K^0]^\top \in \mathbb{R}^{K\times d}.
\end{equation}
This design is causal because $Q_0$ depends only on features and the predicted heatmap from the current forward pass.

The heatmap can also modulate positional encoding as $PE'(i,j)=PE(i,j)\odot(1+\gamma H_b(i,j))$, which enhances foreground positions while preserving background context and is more stable than a purely binary mask.

\subsection{Heatmap-Gated Lite Snake Convolution}

\begin{figure}[!t]
    \centering
    \includegraphics[width=\columnwidth,trim=0 0 0 8,clip]{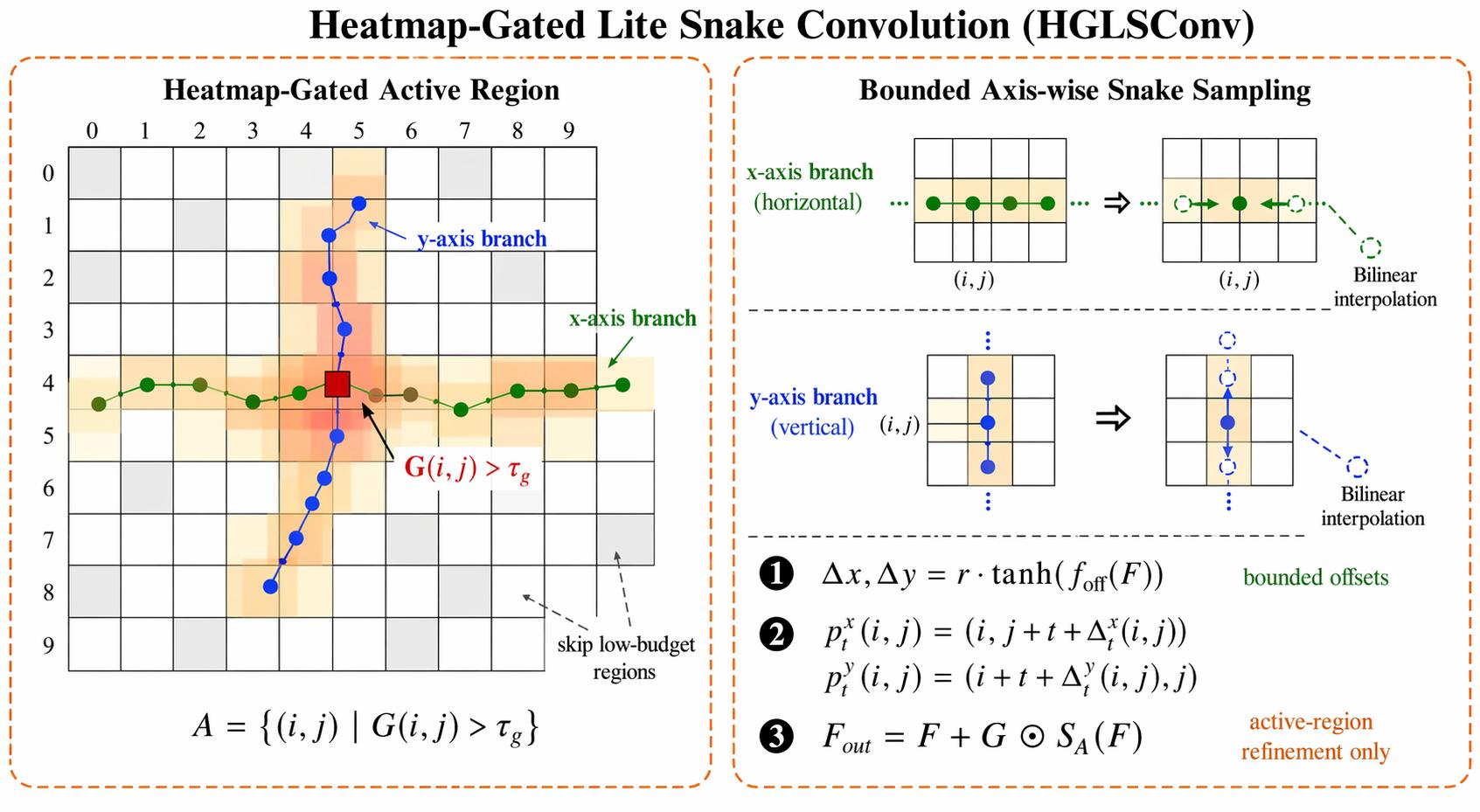}
    \caption{Illustration of Heatmap-Gated Lite Snake Convolution (HGLSConv). HGLSConv activates local refinement only in high-budget regions and uses bounded axis-wise snake sampling for the horizontal and vertical branches.}
    \label{fig:hglsconv}
\end{figure}

Tiny aerial objects usually occupy only a few feature-map cells, so their local shape and boundary evidence can be weakened by downsampling and cluttered backgrounds. Standard convolution is efficient but has fixed sampling locations, while deformable or snake-style refinement can improve local geometry at the cost of dense offset prediction and dense sampling. The motivation of HGLSConv is therefore to preserve the useful local adaptability of snake-style sampling while restricting it to heatmap-indicated small-object candidate regions. HGSQ uses \textbf{Heatmap-Gated Lite Snake Convolution} (HGLSConv), illustrated in Figure~\ref{fig:hglsconv}, which performs local geometric refinement only where the heatmap indicates possible small objects.

Before introducing HGLSConv, we briefly revisit deformable convolution. It improves geometric adaptability by sampling shifted positions with learned offsets, but has two limitations for real-time aerial small object detection: offsets are usually predicted densely over all spatial locations, and unconstrained two-dimensional offsets may scatter sampling points around tiny objects.

HGLSConv keeps the useful local adaptability of deformable sampling but makes it more suitable for real-time small-object refinement. It differs from standard deformable convolution in three aspects: (1) the sampling path is axis-wise rather than fully two-dimensional; (2) the offsets are bounded to avoid uncontrolled deformation; and (3) the snake refinement is computed only on heatmap-activated positions instead of the whole feature map. As shown in Figure~\ref{fig:hglsconv}, this design separates the heatmap-gated active region from the bounded horizontal and vertical sampling branches.

HGLSConv first computes lightweight axis-wise snake features using horizontal and vertical depth-wise convolution branches. The offsets are generated by a lightweight predictor and bounded with a $\tanh$ function, preventing uncontrolled deformation.

For horizontal sampling centered at position $(i,j)$ and kernel index $t\in\{-a,\ldots,a\}$ with $k=2a+1$, the sampled coordinate is
\begin{equation}
    p^x_t(i,j)=\left(i,\; j+t+\Delta^x_t(i,j)\right).
    \label{eq:x_sampling}
\end{equation}
For vertical sampling,
\begin{equation}
    p^y_t(i,j)=\left(i+t+\Delta^y_t(i,j),\; j\right).
    \label{eq:y_sampling}
\end{equation}
Feature values at non-integer coordinates are obtained by bilinear interpolation, and the two axis-wise branches are concatenated and fused by a $1\times1$ projection.

The heatmap gate is then applied by resizing $H_b$ to the feature resolution as $G\in[0,1]^{1\times H\times W}$ and selecting active positions $\mathcal{A}=\{(i,j)\mid G(i,j)>\tau_g\}$. Snake sampling and axis-wise refinement are executed only for $\mathcal{A}$, and the refined responses $S_{\mathcal{A}}(F)\in\mathbb{R}^{C\times H\times W}$ are scattered back to the original feature layout. In deployment, this active-position refinement can be implemented with masked indexing and scatter operations, so the branch is skipped for low-budget background positions. The output is $F_{out}=F+G\odot S_{\mathcal{A}}(F)$, where $G$ is broadcast along the channel dimension, so background regions keep the original lightweight features while high-heatmap regions receive local snake refinement. Therefore, the dominant sampling cost scales with $|\mathcal{A}|$ rather than the full spatial size $HW$, which is suitable for real-time aerial scenes where small objects occupy sparse regions.

During training, a smooth offset regularization term can be used to encourage continuous local sampling instead of scattered offsets.

\subsection{Adaptive Query-Decoder Budgeting}

Different aerial images contain different object densities. A fixed query number can waste computation on sparse images or miss objects in dense images. AQDB estimates image-level density from the heatmap:
\begin{equation}
    \rho=\frac{1}{HW}\sum_{i=1}^{H}\sum_{j=1}^{W}\mathbb{1}\left(H_b(i,j)>\tau\right),
    \label{eq:density}
\end{equation}
where $\tau$ is a heatmap threshold. The query budget is
\begin{equation}
    K(\rho)=\operatorname{clip}\left(K_{min}+\left\lfloor \alpha\rho HW \right\rfloor,\; K_{min},\; K_{max}\right),
    \label{eq:adaptive_k}
\end{equation}
where $K_{min}$ and $K_{max}$ are lower and upper query bounds, and $\alpha$ controls density-to-query scaling.

The maximum decoder depth $D(\rho)$ is also selected from predefined depths $D_{min}$, $D_{mid}$, and $D_{max}$ according to density thresholds. In practice, the decoder can be implemented with a maximum depth $D_{max}$ and an early-exit rule controlled by $D(\rho)$. Sparse images use fewer decoder layers, while dense images retain enough refinement steps.

\subsection{Training Objective and Inference Procedure}

The overall training objective combines the standard detection loss, heatmap supervision, and the optional HGLSConv offset regularization term. If offset regularization is not used, its weight is set to zero.

The inference procedure, summarized in Algorithm~\ref{alg:hgsq}, is single-forward and deterministic:
\begin{algorithm}[t]
    \caption{Single-forward inference of HGSQ}
    \label{alg:hgsq}
    \footnotesize
    \begin{tabular}{@{}p{0.15\columnwidth}p{0.78\columnwidth}@{}}
        \textbf{Input:} & Image $I$; query bounds $K_{min},K_{max}$; decoder depths $D_{min},D_{mid},D_{max}$ \\
        \textbf{Output:} & Detections $\mathcal{D}=\{b_k,c_k,s_k\}_{k=1}^{N}$
    \end{tabular}

    \vspace{0.10cm}
    \begin{tabular}{@{}r@{.\quad}p{0.84\columnwidth}@{}}
        1 & $\{F_l\}_{l=1}^{L} \gets \Phi(I)$ \\
        2 & $H_b \gets \operatorname{HBP}(\{F_l\}_{l=1}^{L})$ \\
        3 & $\rho \gets \frac{1}{HW}\sum_{i,j}\mathbb{1}(H_b(i,j)>\tau)$ \\
        4 & $K \gets K(\rho),\quad D \gets D(\rho)$ \\
        5 & $\widetilde{F}_l \gets \operatorname{HGLSConv}(F_l,H_b)$ for selected $l$ \\
        6 & $\mathcal{P}_K \gets \operatorname{TopK}(H_b,K)$ \\
        7 & $Q_0 \gets \operatorname{HSQS}(\widetilde{F},\mathcal{P}_K,H_b)$ \\
        8 & \textbf{for} $d=1$ \textbf{to} $D$ \textbf{do} \\
        9 & \quad $Q_d \gets \operatorname{DecoderLayer}_d(Q_{d-1},\widetilde{F})$ \\
        10 & \textbf{end for} \\
        11 & $\mathcal{D} \gets \operatorname{DetectionHead}(Q_D)$ \\
        12 & \textbf{return} $\mathcal{D}$ \\
    \end{tabular}
\end{algorithm}

The computational motivation is straightforward. For a decoder with $K$ queries and $D$ layers, the dominant cross-attention and feed-forward costs scale approximately linearly with $D$ and $K$, while self-attention also benefits from reducing $K$. By replacing fixed dense queries with $K(\rho)$ heatmap-selected queries and using $D(\rho)$ adaptive depth, HGSQ reduces background decoding while preserving computation for dense small-object scenes.

\section{Experiments}

This section follows the real-time signal, image, and video processing scope of the target journal. The experiments verify three claims. First, HGSQ improves aerial small-object accuracy while preserving real-time throughput. Second, the improvement comes from the proposed computation-allocation modules rather than from a stronger backbone or a different training setting. Third, the reported efficiency is reflected by measured wall-clock latency, not only by theoretical GFLOPs. Accordingly, Table~\ref{tab:main_results} evaluates the overall accuracy-speed trade-off, Table~\ref{tab:ablation_fairness} isolates the contribution of each module, and Table~\ref{tab:latency_breakdown} verifies the end-to-end deployment cost. The re-evaluated runtime results follow the stated experimental protocol, while other published detectors are included as reference points when identical TensorRT reproduction is unavailable.

\subsection{Real-Time Evaluation Protocol}

Latency is measured with batch size 1 and input size $640\times640$ on an NVIDIA RTX 4070 using PyTorch 1.13.1, CUDA 12.1, and TensorRT FP16 with model fusion when supported. Each speed result is averaged over 1000 test iterations after 200 warm-up iterations, and the reported latency includes preprocessing, model inference, and postprocessing. For the most directly comparable Transformer-based and same-backbone baselines, RT-DETR-R50, RT-DETR-HGNetv2-X, the HGNetv2 baseline, HMPE, and HGSQ are re-evaluated under the same RTX 4070 TensorRT FP16 environment. The reported GFLOPs and latency of HGSQ include the HBP branch, HSQS, HGLSConv, AQDB, decoder, and detection head. Batch size 1 is used because aerial inspection, UAV perception, and edge deployment usually process incoming frames sequentially rather than as large offline batches.

We report both algorithmic complexity and measured wall-clock latency, because GFLOPs alone does not fully reflect real-time performance on deployment hardware. In particular, sparse query selection, active-position refinement, and adaptive decoder depth may change memory access and operator scheduling in ways that are not fully captured by multiply-add counts. Therefore, the evaluation reports Params, GFLOPs, latency, and FPS together.

\subsection{Main Comparison on Aerial Small Object Datasets}

The main comparison focuses on NWPU VHR-10 and VisDrone2019. NWPU VHR-10 evaluates high-resolution remote-sensing objects, while VisDrone2019 provides UAV small-object scenes with denser targets, larger background regions, and stronger real-time motivation. These two datasets test complementary aspects of the proposed method: NWPU evaluates whether HGSQ preserves high-resolution localization quality, whereas VisDrone evaluates whether sparse heatmap-guided computation remains effective in dense UAV scenes. We therefore focus on dedicated aerial and UAV benchmarks rather than general object detection datasets.

The compared methods include efficient YOLO detectors, real-time DETR variants, and recent UAV or remote-sensing small-object detectors. This comparison is intended to evaluate the accuracy-speed trade-off rather than only the maximum FPS. Extremely lightweight models may achieve high frame rates, but they often lose small-object accuracy. Conversely, heavy Transformer or deformable-refinement models may improve accuracy but violate real-time latency constraints. HGSQ operates in the high-accuracy real-time region by reducing background queries and allocating local refinement only to heatmap-activated positions.

\begin{table*}[!t]
\centering
\caption{Main comparison on NWPU VHR-10 and VisDrone2019. Methods marked with $^\dagger$ are re-evaluated under our RTX 4070 TensorRT FP16 protocol. Other published detectors are included as reference points for the accuracy-speed trade-off when identical TensorRT reproduction is unavailable.}
\label{tab:main_results}
\renewcommand{\arraystretch}{1.00}
\setlength{\extrarowheight}{0pt}
\fontsize{6.0}{6.6}\selectfont
\setlength{\tabcolsep}{0.6pt}
\resizebox{\textwidth}{!}{%
\begin{tabular}{@{}l l c c c c c c c c@{}}
\toprule
\multirow{3}{*}{Method} & \multirow{3}{*}{Backbone / Scale} & \multicolumn{5}{c}{Datasets} & \multicolumn{3}{c}{Efficiency} \\
\cmidrule(l{1pt}r{1pt}){3-7}\cmidrule(l{1pt}r{1pt}){8-10}
& & \multicolumn{2}{c}{NWPU VHR-10} & \multicolumn{3}{c}{VisDrone2019} & \multirow{2}{*}{Params(M)} & \multirow{2}{*}{GFLOPs} & \multirow{2}{*}{FPS} \\
\cmidrule(l{1pt}r{1pt}){3-4}\cmidrule(l{1pt}r{1pt}){5-7}
& & mAP50 & mAP50:95 & mAP50 & mAP50:95 & AP-small & & & \\
\midrule
YOLOv8s & CSPDarknet-s & 91.80 & 64.70 & 43.0 & 26.0 & 14.6 & 11.1 & 28.5 & 123.0 \\
YOLOv11s & C3k2/CSP-s & 92.30 & 66.00 & 41.6 & 25.2 & 13.9 & 9.4 & 21.3 & 128.0 \\
YOLOv11m & C3k2/CSP-m & 92.68 & 68.89 & 47.3 & 29.3 & 18.4 & 20.0 & 67.7 & 108.0 \\
SOD-YOLO & YOLOv8-m based & 93.80 & 67.80 & 52.6 & 35.1 & 23.7 & 22.6 & 94.9 & 72.5 \\
RT-DETR-R50$^\dagger$ & ResNet50 & 92.60 & 60.26 & 50.3 & 32.0 & 20.7 & 42.0 & 136.0 & 108.0 \\
RT-DETR-HGNetv2-X$^\dagger$ & HGNetv2-X & 93.60 & 64.60 & 52.4 & 34.2 & 22.6 & 67.0 & 234.0 & 74.0 \\
UAV-DETR & CasNet / RT-DETR based & 93.40 & 65.20 & 51.6 & 33.8 & 22.8 & 16.8 & 49.9 & 83.3 \\
Drone-DETR & RT-DETR enhanced & 94.20 & 66.80 & 53.9 & 35.4 & 24.1 & 28.7 & 128.3 & 30.0 \\
Baseline$^\dagger$ & HGNetv2 & 92.60 & 60.20 & 48.2 & 30.8 & 19.7 & 76.8 & 62.4 & 82.0 \\
HMPE$^\dagger$ & HGNetv2 & 94.50 & 67.20 & 53.6 & 35.2 & 24.0 & 77.3 & 57.0 & 84.0 \\
\rowcolor{LightGreen}
HGSQ, ours$^\dagger$ & HGNetv2 & 95.10 & 68.10 & 54.8 & 36.2 & 25.6 & 76.4 & 48.6 & 96.0 \\
\bottomrule
\end{tabular}%
}
\end{table*}

Table~\ref{tab:main_results} shows that HGSQ achieves the best accuracy among the compared methods while maintaining real-time throughput. On NWPU VHR-10, HGSQ improves mAP50 and mAP50:95 over the baseline, which indicates that sparse query selection does not remove useful small-object evidence. On VisDrone2019, HGSQ also obtains the highest mAP50, mAP50:95, and AP-small. This is important because VisDrone contains many tiny objects under complex UAV viewpoints, where dense background tokens can dominate the decoder budget.

Compared with HMPE, HGSQ increases FPS from 84.0 to 96.0 while improving VisDrone AP-small from 24.0 to 25.6. The simultaneous improvement in AP-small and FPS supports the central claim of this paper: heatmap guidance is more useful when it is used as a computation budget map instead of a post-hoc visualization signal. Compared with RT-DETR-HGNetv2-X and Drone-DETR, HGSQ uses fewer GFLOPs and reaches higher FPS, showing that the gain is not obtained by simply increasing model capacity. Compared with YOLOv8s and YOLOv11s, HGSQ is slower but substantially more accurate on aerial small objects. Therefore, the main advantage of HGSQ is a stronger accuracy-speed trade-off in the high-accuracy real-time regime.

\subsection{Ablation and Same-Backbone Fairness Study}

A common concern in detector comparison is that the reported gain may come from a stronger backbone rather than from the proposed modules. Table~\ref{tab:ablation_fairness} therefore uses the same HGNetv2 backbone and training setting for the core variants. The comparison starts from an RT-DETR-style baseline, then changes the decoder depth and progressively adds HBP, HSQS, HGLSConv, and AQDB. This design isolates whether HGSQ improves the detector by better allocating computation, rather than by changing the feature extractor.

\begin{table*}[!t]
\centering
\caption{Ablation and same-backbone fairness study under the HGNetv2 backbone and identical training setting.}
\label{tab:ablation_fairness}
\renewcommand{\arraystretch}{1.00}
\setlength{\extrarowheight}{0pt}
\fontsize{6.0}{6.6}\selectfont
\setlength{\tabcolsep}{0.6pt}
\resizebox{\textwidth}{!}{%
\begin{tabular}{@{}l l c c c c c c c c c@{}}
\toprule
\multirow{2}{*}{Variant} & \multirow{2}{*}{Backbone} & \multirow{2}{*}{Decoder} & \multicolumn{4}{c}{Modules} & \multicolumn{2}{c}{Accuracy} & \multicolumn{2}{c}{Efficiency} \\
\cmidrule(l{1pt}r{1pt}){4-7}\cmidrule(l{1pt}r{1pt}){8-9}\cmidrule(l{1pt}r{1pt}){10-11}
& & & HBP & HSQS & HGLSConv & AQDB & mAP50 & mAP50:95 & GFLOPs & FPS \\
\midrule
RT-DETR-HGNetv2 baseline & HGNetv2 & 6 & -- & -- & -- & -- & 92.90 & 62.10 & -- & 80.0 \\
Baseline & HGNetv2 & original & -- & -- & -- & -- & 92.60 & 60.20 & 62.4 & 82.0 \\
Baseline, 3 decoder & HGNetv2 & 3 & -- & -- & -- & -- & 91.80 & 58.90 & -- & 112.0 \\
+ HBP + HSQS & HGNetv2 & 3 & \checkmark & \checkmark & -- & -- & 93.80 & 64.20 & 51.0 & 108.0 \\
+ HBP + HSQS + HGLSConv & HGNetv2 & 3 & \checkmark & \checkmark & \checkmark & -- & 94.70 & 66.70 & 52.4 & 100.0 \\
+ Adaptive $K$ & HGNetv2 & 3 & \checkmark & \checkmark & \checkmark & adaptive $K$ & 94.90 & 67.40 & 49.3 & 99.0 \\
\rowcolor{LightGreen}
Full HGSQ & HGNetv2 & adaptive ($D_{max}=3$) & \checkmark & \checkmark & \checkmark & \checkmark & 95.10 & 68.10 & 48.6 & 96.0 \\
\bottomrule
\end{tabular}%
}
\end{table*}

Table~\ref{tab:ablation_fairness} shows that shallow decoding alone is not sufficient: reducing the decoder to three layers increases FPS but decreases mAP50:95 from 60.20 to 58.90. Adding heatmap-guided query selection reduces GFLOPs and improves accuracy, and HGLSConv further recovers small-object localization quality. The full HGSQ model reaches 68.10 mAP50:95 and 96.0 FPS under the same backbone, showing that the improvement comes from sparse query selection, heatmap-gated refinement, and adaptive budgeting rather than backbone replacement.

\subsection{Sparse Query and Query Quality Analysis}

Sparse query selection connects heatmap prediction with real-time Transformer decoding. Table~\ref{tab:query_budget} reports the query-budget trade-off, and Table~\ref{tab:query_quality} verifies whether the selected queries are better aligned with foreground objects.

\begin{table}[!t]
\centering
\caption{HSQS query-budget ablation with different query selection strategies.}
\label{tab:query_budget}
\renewcommand{\arraystretch}{1.00}
\setlength{\extrarowheight}{0pt}
\fontsize{6.0}{6.6}\selectfont
\setlength{\tabcolsep}{0.5pt}
\begin{tabular}{l c c c c}
\toprule
\multirow{2}{*}{Query strategy} & \multicolumn{2}{c}{Budget} & \multirow{2}{*}{mAP50:95} & \multirow{2}{*}{FPS} \\
\cmidrule(l{0pt}r{0pt}){2-3}
& $K$ & Layers & & \\
\midrule
Learned dense query & fixed & 6 & 61.50 & 80.0 \\
Encoder top-$K$ query & 300 & 6 & 62.10 & 83.3 \\
Heatmap top-$K$ query & 100 & 3 & 64.00 & 128.0 \\
Heatmap top-$K$ query & 300 & 3 & 67.50 & 99.0 \\
Heatmap top-$K$ query & 500 & 3 & 68.00 & 85.0 \\
\rowcolor{LightGreen}
Adaptive heatmap query & adaptive & $D_{max}=3$ & 68.10 & 96.0 \\
\bottomrule
\end{tabular}
\end{table}

\begin{table}[!t]
\centering
\caption{Query quality analysis for dense, encoder-selected, and heatmap-selected queries.}
\label{tab:query_quality}
\renewcommand{\arraystretch}{1.00}
\setlength{\extrarowheight}{0pt}
\fontsize{6.0}{6.6}\selectfont
\setlength{\tabcolsep}{0.5pt}
\begin{tabular}{l c c c c}
\toprule
\multirow{2}{*}{Method} & \multicolumn{3}{c}{Query quality} & \multirow{2}{*}{Avg. queries} \\
\cmidrule(l{0pt}r{0pt}){2-4}
& FG ratio & GT recall & Duplicate ratio & \\
\midrule
Learned dense query & 18.4 & 82.1 & 31.5 & 300 \\
Encoder top-$K$ query & 24.7 & 86.8 & 27.2 & 300 \\
Heatmap top-$K$ query & 41.3 & 91.6 & 18.5 & 300 \\
\rowcolor{LightGreen}
Adaptive heatmap query & 45.8 & 92.4 & 15.9 & 224 \\
\bottomrule
\end{tabular}
\end{table}

Table~\ref{tab:query_budget} shows that query reduction alone is not enough: $K=100$ gives the highest FPS but loses accuracy, while larger fixed budgets improve mAP50:95 at the cost of speed. Table~\ref{tab:query_quality} further shows that heatmap-selected queries have higher foreground ratios and lower duplicate ratios than learned dense or encoder-selected queries. The adaptive strategy achieves the best mAP50:95 with only 224 average queries, indicating that HGSQ is faster because it removes mostly background or duplicate queries while preserving foreground evidence.

\subsection{HGLSConv Analysis}

HGLSConv is designed to improve local refinement for tiny objects while avoiding dense deformable sampling on background regions. Table~\ref{tab:hglsconv_compare} compares it with standard convolution, dilated convolution, deformable convolution, DSConv, and the previous LSConv design under the same real-time setting.

\begin{table}[!t]
\centering
\caption{Comparison of local refinement modules.}
\label{tab:hglsconv_compare}
\renewcommand{\arraystretch}{1.00}
\setlength{\extrarowheight}{0pt}
\fontsize{5.8}{6.3}\selectfont
\setlength{\tabcolsep}{0.5pt}
\begin{tabular}{@{}l c c c c c c@{}}
\toprule
Module & Gate & mAP50 & mAP50:95 & AP-s & GFLOPs & FPS \\
\midrule
Std. Conv & -- & 93.70 & 64.80 & 21.8 & 50.1 & 104.0 \\
Dilated & -- & 94.00 & 65.50 & 22.6 & 51.0 & 101.0 \\
DCNv2/3 & -- & 94.80 & 67.70 & 24.4 & 61.4 & 72.0 \\
DSConv & -- & 94.20 & 66.10 & 23.1 & 54.8 & 88.0 \\
LSConv & -- & 93.91 & 63.69 & 24.0 & 57.0 & 84.0 \\
Lite Snake & -- & 94.50 & 66.80 & 24.8 & 52.9 & 95.0 \\
\rowcolor{LightGreen}
HGLSConv & \checkmark & 95.10 & 68.10 & 25.6 & 48.6 & 96.0 \\
\bottomrule
\end{tabular}
\vspace{-0.4em}
\end{table}

As shown in Table~\ref{tab:hglsconv_compare}, standard convolution and dilated convolution are efficient but provide weaker AP-small, because their sampling locations are fixed. DCNv2/DCNv3 improves AP-small to 24.4, but the GFLOPs increase to 61.4 and FPS drops to 72.0 due to dense offset prediction and dense deformable sampling. DSConv and the previous LSConv design improve geometric modeling, but they still apply refinement more broadly than necessary for sparse aerial targets.

HGLSConv achieves the highest AP-small and mAP50:95 while keeping 96.0 FPS. Its heatmap gate prevents the refinement branch from being applied to low-budget background regions, so the method retains local shape adaptation without the dense offset-prediction cost of DCNv2/DCNv3. This supports the design choice that HGLSConv is a heatmap-triggered local refinement operator for sparse aerial small-object regions.

\subsection{End-to-End Latency Breakdown}

For a real-time signal, image, and video processing journal, the final efficiency claim should be supported by end-to-end latency instead of GFLOPs alone. Table~\ref{tab:latency_breakdown} decomposes the measured runtime into preprocessing, backbone, neck, heatmap head, local refinement, encoder, decoder, and postprocessing. This breakdown is used to answer where the speed gain comes from and whether the heatmap branch introduces hidden overhead.

\begin{table}[!t]
\centering
\caption{End-to-end latency breakdown measured with batch size 1 and TensorRT FP16 deployment.}
\label{tab:latency_breakdown}
\renewcommand{\arraystretch}{1.03}
\setlength{\extrarowheight}{0pt}
\fontsize{6.0}{6.6}\selectfont
\setlength{\tabcolsep}{0.5pt}
\begin{tabular}{l c c c}
\toprule
\multirow{2}{*}{Component} & \multicolumn{3}{c}{Latency (ms)} \\
\cmidrule(l{0pt}r{0pt}){2-4}
& RT-DETR-HGNetv2-X & HMPE & HGSQ \\
\midrule
Preprocessing & 0.6 & 0.6 & 0.6 \\
Backbone & 4.8 & 4.2 & 4.2 \\
Neck & 1.5 & 1.2 & 1.2 \\
Heatmap head & -- & 0.4 & 0.3 \\
LSConv / HGLSConv & -- & 1.8 & 0.8 \\
Encoder & 2.8 & 2.0 & 1.8 \\
Decoder & 2.6 & 1.0 & 0.8 \\
Postprocessing & 1.2 & 0.7 & 0.7 \\
\midrule
Total latency & 13.5 & 11.9 & 10.4 \\
FPS & 74.0 & 84.0 & 96.0 \\
\bottomrule
\end{tabular}
\end{table}

The latency breakdown in Table~\ref{tab:latency_breakdown} confirms that the real-time gain is not only reflected by GFLOPs. HBP adds only 0.3 ms, which is small compared with the backbone and decoder costs. More importantly, active-position HGLSConv reduces the local refinement latency from 1.8 ms in HMPE to 0.8 ms because snake sampling is executed only on heatmap-activated positions. The encoder and decoder latency are also reduced, which is consistent with the sparse query and adaptive depth design.

The total latency decreases from 11.9 ms in HMPE to 10.4 ms in HGSQ, corresponding to an FPS increase from 84.0 to 96.0. This result supports the journal-oriented claim that HGSQ improves deployable real-time performance, not only theoretical complexity. In addition, the latency table clarifies that the heatmap budget predictor does not create the causal or computational overhead associated with Grad-CAM-style heatmaps. The heatmap is produced in the same forward pass and immediately reused for query selection, local refinement, and adaptive decoding.

To further analyze the effect of AQDB under variable scene densities, we divide VisDrone2019 images according to the number of annotated objects per image: sparse scenes contain no more than 5 objects, medium-density scenes contain 6--15 objects, and dense scenes contain more than 15 objects. Table~\ref{tab:density_latency} reports the corresponding heatmap recall, adaptive query budget, decoder depth, accuracy, and latency distribution. The reported 96.0 FPS is the average throughput over the full test set rather than the worst-case throughput. In dense scenes, AQDB increases the average query budget to 403 and uses the maximum decoder depth, so the average FPS decreases to 70.4 and the worst-case latency reaches 22.7 ms. This confirms that the adaptive budget introduces scene-dependent latency variation. Nevertheless, HGSQ remains real-time under batch-size-one sequential inference, and the overall latency remains 10.4 ms. Additional size-wise and occlusion-wise HBP recall and false-positive analyses are provided in the response letter due to the page limit of the manuscript.

\begin{table}[!t]
\centering
\caption{Density-aware AQDB behavior and latency on VisDrone2019.}
\label{tab:density_latency}
\renewcommand{\arraystretch}{1.00}
\setlength{\extrarowheight}{0pt}
\fontsize{5.8}{6.3}\selectfont
\setlength{\tabcolsep}{0.4pt}
\begin{tabular}{@{}l c c c c c c c@{}}
\toprule
Group & HBP & $K$ & $D$ & AP50 & Lat. & Worst & FPS \\
& Rec. & & & & (ms) & (ms) & \\
\midrule
Sparse & 92.8 & 152 & 1.8 & 57.0 & 9.1 & 13.2 & 109.9 \\
Medium & 91.1 & 242 & 2.3 & 55.0 & 10.5 & 15.4 & 95.2 \\
Dense & 89.3 & 403 & 3.0 & 47.5 & 14.2 & 22.7 & 70.4 \\
\rowcolor{LightGreen}
Overall & 90.6 & 224 & 2.2 & 54.8 & 10.4 & 22.7 & 96.0 \\
\bottomrule
\end{tabular}
\vspace{-0.4em}
\end{table}

\section{Conclusion and Outlook}

This paper presents HGSQ, a heatmap-guided sparse query detector for real-time aerial small object detection. HGSQ uses a single-forward heatmap budget to guide sparse query selection, heatmap-gated local refinement, and adaptive query-decoder budgeting. Experiments on NWPU VHR-10 and VisDrone2019 show that HGSQ improves small-object accuracy while reducing GFLOPs and end-to-end latency, confirming the effectiveness of heatmap-guided computation allocation for real-time aerial detection. The main limitation is that HGSQ still depends on the quality of the predicted heatmap: extremely tiny, low-contrast, or heavily occluded objects may receive weak heatmap responses, and dense scenes can increase the adaptive query budget and latency. Future work will examine broader deployment scenarios and hardware backends while keeping the same single-forward computation-allocation principle.

\section*{Declarations}

\bmhead{Author contributions} Y.Z. performed the conceptualization, methodology, implementation, experiments, analysis, visualization, and manuscript writing.

\bmhead{Funding} This work was supported by the National Natural Science Foundation of China, the Engineering Research Center of Autonomous Unmanned System Technology, Ministry of Education, and the Big Data Computing Center of Southeast University.

\bmhead{Data and code availability} The NWPU VHR-10 and VisDrone2019 datasets are publicly available. Processed data, code, and trained models are available from the author upon reasonable request.

\bmhead{Competing interests} The authors declare no competing interests.

\end{document}